\documentclass{article}
\usepackage{spconf,amsmath,graphicx,hyperref}

\title{FinLLM-KG: Knowledge Graph-Enhanced LLMs for Financial Sentiment Analysis in Algorithmic Trading}
\name{Giorgos Iacovides, Oliver Hannibal, Wuyang Zhou, Danilo Mandic}
\address{Department of Electrical and Electronic Engineering, Imperial College London, UK}
\usepackage{booktabs}
\usepackage{makecell}
\usepackage{graphicx}

\begin{document}
\ninept
\maketitle
\begin{abstract}
Recent advances in Generative AI have substantially improved financial sentiment analysis through post-trained financial large language models (LLMs). However, existing approaches remain confined to predicting the sentiment of each article in isolation and operate as black boxes, which limits contextual understanding and hinders deployment in high-stakes financial decision-making. To this end, we introduce FinLLM-KG, a model-agnostic framework that combines causal LLMs with a financial knowledge graph, in order to inject temporally grounded, company-level knowledge from SEC 10-K filings at inference time, without updating model weights. Within FinLLM-KG, context is supplied either as a natural-language company profile or as a small set of relevance-ranked factual triples. This enables sentiment predictions to account for broader firm-level information rather than the article alone, at almost zero computational cost and for any causal LLM. In addition, attributing each article-level sentiment score to individual sentences and to the retrieved knowledge-graph facts makes the evidence behind the prediction inspectable, thereby enhancing explainability and supporting human-in-the-loop review. Simulations demonstrate that FinLLM-KG consistently improves portfolio performance over the state-of-the-art baseline.
\end{abstract}
\section{Introduction}
\label{sec:intro}
Financial sentiment analysis (FSA) refers to the quantification of opinions expressed in news and other sources of textual financial data and aims to categorize whether the overall sentiment is positive, negative, or neutral — the so-called valence. The rise of Generative AI (GenAI), and in particular large language models (LLMs), has made FSA an increasingly important component of algorithmic and event-driven trading strategies, owing to their ability to process and interpret vast amounts of unstructured data. 
The ability to effectively leverage these technologies to extract actionable insights from complex financial narratives, at scale, enables the identification of aggregate trends and information that may otherwise be difficult to capture. Consequently, this provides an information advantage for downstream financial applications such as stock price prediction \cite{stock_prices} and algorithmic trading \cite{trad_strats, finllama,findpo}. \par 
Despite conceptual benefits, the diverse, nuanced, and domain-specific nature of financial text makes reliable sentiment extraction challenging. Post-trained LLMs \cite{finllama, findpo,fingpt, zhang2023instructfingpt,finsmart}, specifically adapted to the language of financial markets, have recently emerged as a promising approach to FSA, due to their ability to capture complex financial semantics without incurring the computational cost of pre-training LLMs from scratch. Despite advances, current sentiment-driven trading systems exhibit two fundamental limitations. \par  
First, they analyze articles in isolation, ignoring the interconnected structure of financial markets, where news about one firm can rapidly affect its competitors, suppliers, customers, and other economic partners. Second, they largely operate as black-box systems, producing sentiment predictions without explicitly identifying the text that contributed to them — a critical limitation in finance, where explainability and human-in-the-loop oversight are essential.\par 
\begin{figure*}[t] 
  \centering
  \includegraphics[width=0.96\textwidth]{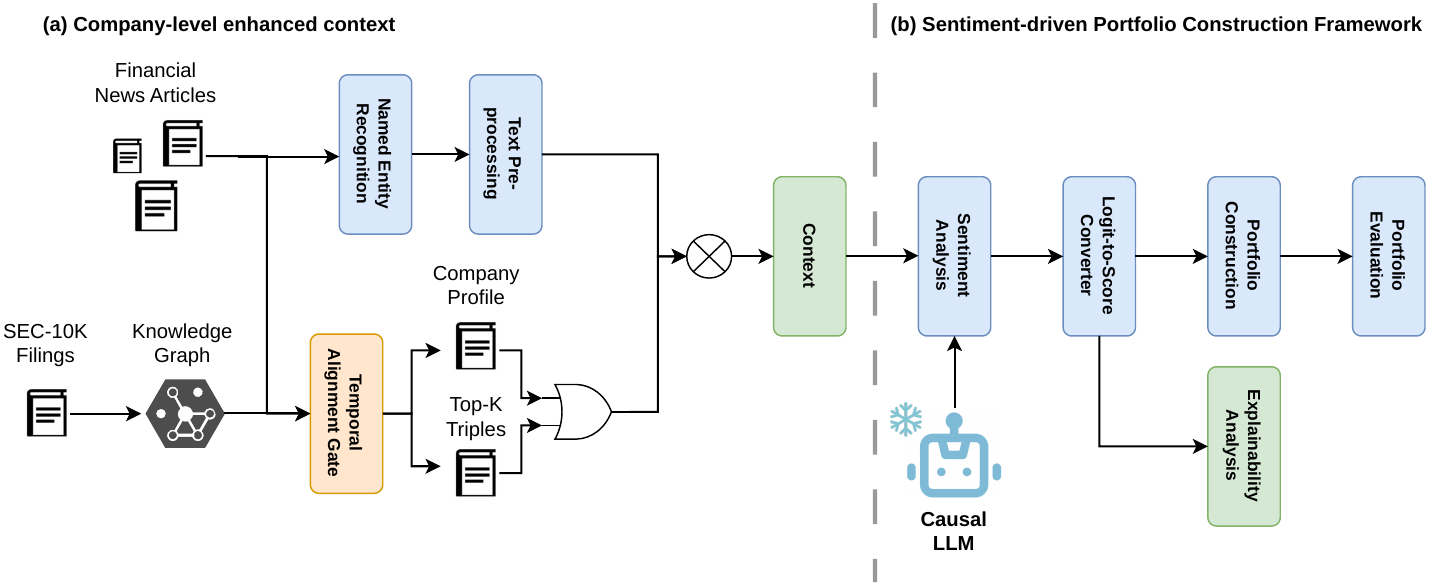}
  \caption{Overview of FinLLM-KG. (a) Financial news articles are tagged and preprocessed. In parallel, fixed-relation 10-K triples from FinReflectKG \cite{finreflectkg} are restricted by a temporal alignment gate to the latest 10-K filed strictly before the article. Eligible facts are supplied either as a company profile or as the top-$k$ retrieved triples and mixed with the article; in the baseline, the context is the article alone. (b) A frozen causal LLM performs sentiment analysis on the resulting context (prompt). The sentiment signal is used in the portfolio construction framework and, for the proposed methods only, can be decomposed into sentence-level and triple-level attributions for explainability.}
  \label{fig:pipeline}
\end{figure*}
Simultaneously, knowledge graphs (KGs) have emerged as a powerful paradigm for explicitly representing financial information. Unlike LLMs, which encode knowledge implicitly in their parameters, KGs represent entities and relationships as structured, inspectable triples, each encoding a factual relationship between two entities. In finance, this is particularly suited to information extracted from SEC 10-K filings \cite{sec_10k,finreflectkg}, where factual claims can be grounded in timestamped public disclosures and traced to their sources. Importantly, LLMs and KGs are complementary: LLMs provide semantic understanding of complex financial narratives, while KGs provide explicit relational structure and provenance. Combining the two offers a pathway beyond isolated sentiment classification towards more context-aware and explainable framework for financial sentiment analysis. It is therefore natural to ask:
\begin{itemize}
    \item Can knowledge graphs be used in conjunction with finance-specific LLMs to enhance the predictive accuracy and explainability of sentiment-driven trading strategies?
\end{itemize}
To this end, we develop FinLLM-KG, a framework that combines: i) an audited knowledge graph of facts extracted from 10-K filings, and ii) finance-specific causal LLMs, to improve the predictive power and explainability of sentiment-based alpha generation. In particular, FinLLM-KG injects explicit company-level context into the signal-generation stage at inference time, while retaining the generative capabilities of the underlying LLM. Within this framework, each article is scored alongside information drawn from the latest publicly available 10-K filing, supplied either as a natural-language company profile or as a small set of relevance-ranked factual triples (e.g., \textit{NVIDIA --- PARTNERS\_WITH --- Microsoft}). In this way, sentiment predictions can account for the broader network of relationships surrounding a firm, while the facts underlying each score remain explicit and inspectable rather than implicit in the model parameters.  The main contributions of this work are as follows. 
\begin{itemize}
\item We develop FinLLM-KG, a knowledge graph-enhanced framework that integrates audited 10-K facts with finance-specific causal LLMs for financial sentiment analysis. 
\item We introduce an inference-time contextualization mechanism that augments financial narratives with company profiles and relevance-ranked factual relations, enabling more context-aware and explainable sentiment predictions.
\item Extensive experiments demonstrate that FinLLM-KG improves predictive and trading performance while providing explicit, inspectable evidence underlying its predictions.
\end{itemize}
\section{Related Work}
\noindent \textbf{Sentiment Analysis with Large Language Models.} 
Recent work has explored several post-training paradigms for financial sentiment analysis, including supervised fine-tuning (SFT) of classification LLMs \cite{finllama}, causal SFT of generative LLMs \cite{fingpt,zhang2023instructfingpt}, preference optimisation through DPO \cite{findpo}, and market-aligned reinforcement learning (RL) \cite{finsmart}. Despite their different objectives, these approaches share two key limitations: they generally analyse articles in isolation and provide predictions without revealing the supporting textual or contextual evidence. This black-box behaviour restricts human oversight in high-stakes financial decisions. FinLLM-KG addresses these limitations by integrating company knowledge and providing sentence- and triple-level attributions for each prediction. \newline
\textbf{Knowledge Graphs in Finance.} 
Knowledge graphs (KGs) have been increasingly used to represent structured financial knowledge from sources including financial news and corporate filings. Notable examples include event-centric graphs such as FinDKG \cite{findkg} and filing-derived graphs such as FinReflectKG \cite{finreflectkg}, which extracts schema-constrained triples from SEC 10-K filings using an LLM-based critic--corrector framework. Financial KGs support question answering, retrieval, temporal analysis, and forecasting \cite{kg_llm_survey,kg_llm_integration,financial_kg_forecasting}, and have been integrated with LLMs through retrieval augmentation or joint modelling \cite{kg_llm_survey,kg_llm_roadmap}. In this work, we use FinReflectKG to augment a finance-specific causal LLM with temporally grounded company relationships from corporate filings.
\section{FINLLM-KG}
Our work is designed to enhance the contextual understanding and explainability of causal LLMs for financial sentiment analysis by augmenting articles with company-level knowledge from a financial knowledge graph, rather than evaluating them in isolation. The approach is model-agnostic and requires no retraining, using the graph as external company memory at inference time. This broader firm-level context aims to improve sentiment accuracy and downstream trading performance. Sentence-level attributions, together with temporally grounded graph evidence, make predictions more transparent and suitable for human-in-the-loop applications. Figure \ref{fig:pipeline} illustrates the overall framework.
\subsection{Context Construction} \label{context_construction}
FinLLM-KG does not extract a knowledge graph from raw filings. Instead, it adopts FinReflectKG, a published graph of schema-constrained triples derived from SEC 10-K filings for more than 700 companies over fiscal years 2014--2024, constructed using an LLM-based critic--corrector loop \cite{finreflectkg}. Each triple is a $(\textit{head},\textit{relation},\textit{tail})$ fact tagged with a ticker and fiscal year. SEC 10-Ks are a natural source for this task: they are legally mandated, standardised, and dated, providing attributable and temporally ordered financial information \cite{sec_10k}. The underlying filings are obtained through the SEC's EDGAR system. The resulting triples are stored in Neo4j, a graph database, and retrieved at inference time as external, company-specific memory. This company memory is provided to the sentiment LLM through two alternative representations: (i) a natural-language company profile and (ii) relevance-ranked factual triples. Given that a fiscal year is an interval rather than a timestamp, each triple is aligned to the EDGAR filing date of the corresponding 10-K. At scoring time, both forms of context are restricted to the latest 10-K filed \textit{strictly before} the article date. If no such filing exists, no graph context is supplied.
\subsubsection{Company Profiles}
The first representation compresses a firm's triples into a short natural-language profile, allowing the sentiment model to interpret an article in the context of the company's broader situation rather than in isolation. For each ticker and fiscal year, triples are verbalised as $(\textit{head},\textit{relation},\textit{tail})$ sentences, with underscores in relation names replaced by spaces, and passed to Claude Haiku 4.5. Profiles are built cumulatively: the first available year is generated from scratch under five fixed headings (Business, Financials, Momentum, Strategy, Key Risks), while each subsequent year updates the previous profile with that year's new facts, allowing multi-year trends to be stated rather than inferred from a single filing. Length is capped at 500 words for the first year and 700 words thereafter to keep the resulting prompt focused on the article. A year with no new triples carries the previous profile forward. Each completed profile is labelled with its 10-K filing date.
\subsubsection{Relevance-Ranked Triples}
The second representation retains the graph in fact form and, for each article, retrieves only the triples most relevant to its text. Offline, every unique triple for a ticker is verbalised as above, embedded using \texttt{all-MiniLM-L12-v2} \cite{minillm}, and stored in a per-ticker vector index together with its fiscal year. Online, retrieval is confined to the latest 10-K filed \textit{strictly before} the article date, after removing generic relations. The article is embedded with the same encoder, and the top-$k$ triples are retrieved by cosine similarity between the resulting unit-normalised embeddings with the default set to $k=4$.
\subsection{Inference-time Scoring}
As the FinLLM-KG framework operates with a frozen causal language model, its weights are never updated, making the framework model-agnostic and applicable to any causal LLM. In our evaluations, we compare three different conditions in the context supplied to the prompt. The baseline condition supplies the article text alone. The two knowledge-enhanced conditions prepend the company-level context generated in Section \ref{context_construction}---either the latest eligible 10-K profile or the top-$k$ retrieved triples---stamped with its filing provenance. The model is explicitly instructed to judge the article using the background context only as guidance, allowing the current news to take precedence if the two conflict. To strictly enforce the temporal gate, if no filing is yet public for a company at the time of an article, the knowledge-enhanced score is left undefined rather than silently falling back to the baseline.

Given that FinLLM-KG is built upon a causal LLM, its outputs are naturally limited to discrete sentiment labels (i.e., positive, negative, or neutral). To obtain continuous sentiment scores suitable for portfolio construction, we employed the logit-to-score converter proposed in \cite{findpo}, which leverages the internal output logits of the model to quantify the strength of each sentiment prediction without requiring any architectural modifications. Each article is scored via a single forward pass per condition, and this continuous score serves as the signal for portfolio construction and the target of the explainability decompositions.
\section{Sentiment-Driven Portfolio Construction Framework}
\textbf{Data Collection.} Both textual and market data were collected to support the construction of long--short (L/S) portfolios. The textual dataset comprises $\approx$200,000 financial news articles published between February 2015 and June 2021. For the same period, we collected daily stock returns for the 100 companies in our investable universe (S\&P 100) from Yahoo Finance, resulting in 1,672 days of return data per company. \newline
\textbf{Named Entity Recognition (NER).} Since financial articles frequently discuss multiple companies, naively linking an article to a single stock can introduce substantial misalignment into the sentiment signal. To address this, we applied Named Entity Recognition (NER) to identify the organisational entities referenced within each article. Specifically, we employed the BERT-base-NER model~\cite{bert_base_ner}, which recognises four entity types---organisation, person, location, and miscellaneous---and produces a confidence score for each detected entity. Articles were retained only when the confidence score of the target company exceeded 98\%; otherwise, they were discarded. This reduced the corpus of financial articles to $\approx$ 160,000. \newline 
\textbf{Sentiment Analysis.} To assess the financial value of the proposed knowledge-enhanced context, FinLLM-KG was compared with the  FinDPO configuration, a finance-specific causal LLM shown to achieve state-of-the-art portfolio-construction performance \cite{findpo}. Since both configurations use the same sentiment-scoring procedure, this comparison isolates the contribution of company-level knowledge-graph context. Every article within each corpus was evaluated, with sentiment averaged when multiple articles were published for the same company on a given day. \newline
\textbf{Portfolio Construction.} Once the sentiment score for each method was defined for every company, a daily long--short portfolio was constructed. We used the sentiment strength as a parameter to determine which companies should be in a long or a short position. The portfolio construction procedure is summarized below:
\begin{itemize}
\item \textit{Define the Investable Universe.} Although the benchmark universe is based on S\&P 100 constituents, the investable universe is constructed in a point-in-time and dynamic manner. On each trading day, only companies with point-in-time financial news and corresponding market data were included. 

\item \textit{Define the long and short positions.} The sentiment signals produced by each method were used to construct independent portfolios. On each trading day, companies were ranked according to their sentiment score. Companies that did not have sentiment data on a particular day were omitted from the ranking. As the daily sentiment score for each company ranges between -1 and 1, companies with the highest positive sentiment were placed in a long position, while those with the strongest negative sentiment were placed in a short position. 
\item \textit{Portfolio allocation.} Following \cite{finllama,findpo,finsmart}, an equally weighted 35\% long--short portfolio was employed, with the top 35\% of ranked companies allocated to long positions and the bottom 35\% allocated to short positions. 

\item \textit{Determine daily returns.} To ensure a realistic trading setting and eliminate look-ahead bias, sentiment generated from articles published on trading day $t$ was used to construct the portfolio at the market open of the next trading day $t+1$. Positions were then held for one trading day and closed at the subsequent market open, implying that all returns were computed on a next-day open-to-open basis. For each particular day, the number of companies that were held in either a long position or a short position were equal. Consequently, the total portfolio return on a particular day is the difference between the daily long return and the daily short return.
\end{itemize}
\textbf{Portfolio Evaluation.} The performance of the portfolios constructed using FinLLM-KG was assessed against the FinDPO baseline. We considered (i) \textit{Profitability}, measured via cumulative returns, and annualized returns and (ii) \textit{Risk-adjusted performance}, evaluated using the annualized Sharpe ratio, Sortino ratio, and Calmar ratio. In line with the literature, and given the persistently low yields during the sample period \cite{low_yield}, we set the risk-free rate, $R_f=0$.
\section{Experimental Results}
\subsection{Evaluation via Real-World Financial Metrics}
\begin{figure}[t]
\centering
\includegraphics[width=\linewidth]{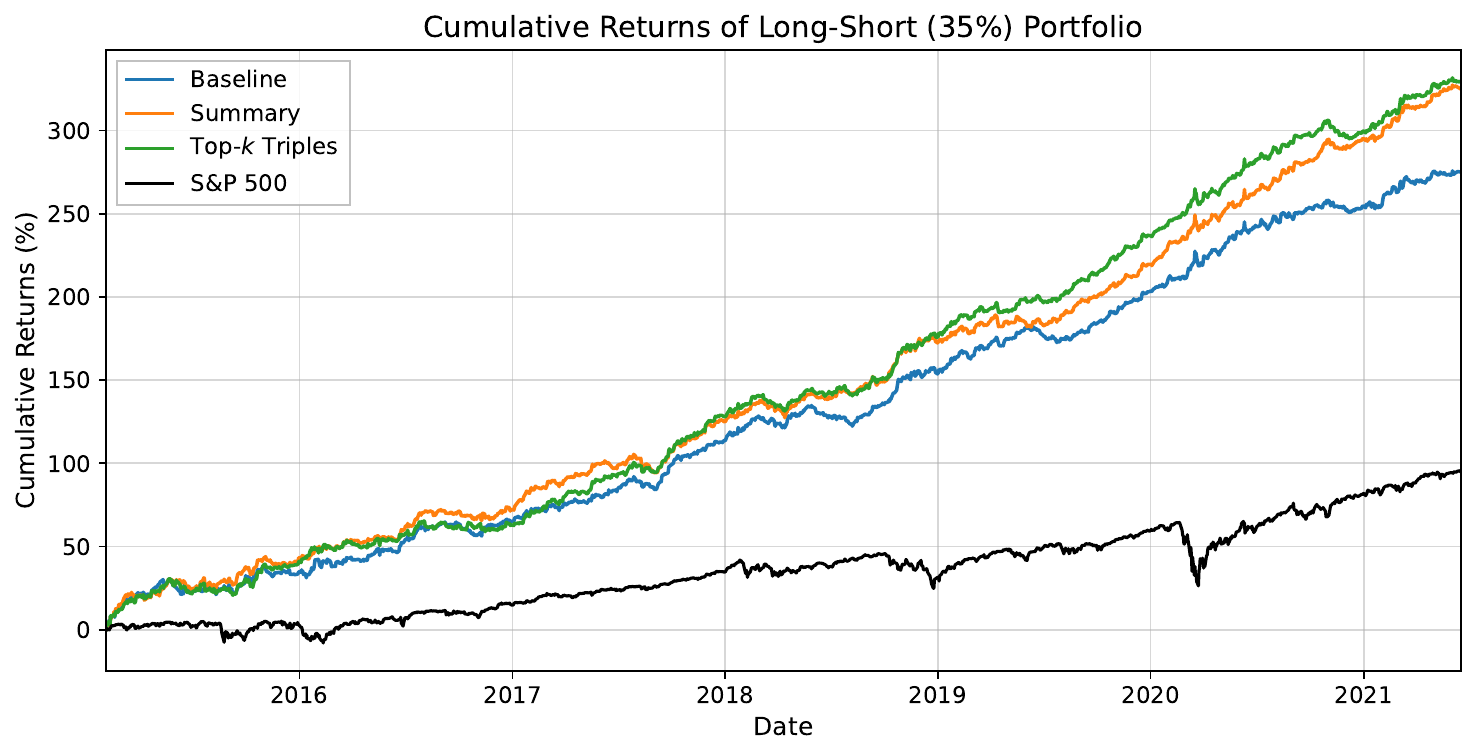}
\caption{Cumulative returns of sentiment-based 35\% long--short portfolios under three conditions: Baseline (article text only), Summary (article text with the latest 10-K company profile), and Top-$k$ Triples (article text with the top-$k$ relevance-ranked factual relations), compared against the S\&P 500 benchmark.}

\label{fig:cumret_0bps}
\end{figure}

The results for the three sentiment-based configurations are illustrated in Figure~\ref{fig:cumret_0bps} and Table~\ref{tab:performance_0bps}. Despite using the same base model as FinDPO \cite{findpo}, which has achieved state-of-the-art performance in portfolio construction, FinLLM-KG improves upon the baseline across all evaluation metrics by incorporating company-level knowledge rather than scoring each article in isolation. 
\begin{table}[h!]
\centering
\caption{Performance metrics of the sentiment-based portfolios at no transaction cost. For each metric, the best performance is shown in bold, and the second-best is underlined.}
\label{tab:performance_0bps}
\resizebox{\linewidth}{!}{%
\begin{tabular}{lccccc}
\specialrule{.1em}{0.2em}{0.3em}
\textbf{Method} & \makecell{\textbf{Cumulative}\\\textbf{Return (\%)}} & \makecell{\textbf{Annualized}\\\textbf{Return (\%)}} & \textbf{Sharpe} & \textbf{Sortino} & \textbf{Calmar} \\
\specialrule{.1em}{0.2em}{0.3em}
\textbf{S\&P 500} & 94.75 & 12.77 & 0.72 & 0.90 & 0.46 \\
\specialrule{.1em}{0.2em}{0.3em}
Baseline & 275.04 & 40.39 & 2.55 & 4.12 & 4.38 \\
\textbf{Summary (Ours)} & \underline{325.06} & \underline{47.98} & \underline{3.07} & \underline{4.60} & \underline{6.07} \\
\textbf{Top-$k$ Triples (Ours)} & \textbf{329.54} & \textbf{48.69} & \textbf{3.15} & \textbf{4.96} & \textbf{6.12} \\
\specialrule{.1em}{0.2em}{0.3em}
\end{tabular}
}
\end{table}\par
The summary and top-$k$ triple configurations outperform the baseline, increasing cumulative returns from 275.04\% to 325.06\% and 329.54\%, respectively. They also improve the Sharpe (2.55 to 3.07/3.15), Sortino (4.12 to 4.60/4.96), and Calmar ratios (4.38 to 6.07/6.12), indicating stronger risk-adjusted performance and improved downside-risk control. As the long--short portfolios are dollar-neutral by construction (SPY betas of $-0.04$ to $-0.06$), these returns are driven by signal alpha rather than market exposure. \par 
Overall, these results indicate that augmenting article-level sentiment analysis with temporally grounded company context produces more informative trading signals than financial news in isolation. The superior top-$k$ results further suggests that compact, article-specific facts offer greater utility than broader company summaries, by enabling the model to focus on the most relevant information at the given point in time.

\subsection{Explainability Analysis}
\label{sec:explainability_results}

In addition to evaluating financial performance, we examined whether FinLLM-KG provides an interpretable account of the sentiment signals used for portfolio construction. Each article was scored as a complete document before attribution methods decompose the resulting score. The explanations therefore identify how individual sentences contribute to the full-article prediction rather than aggregating independently generated sentence scores.

\begin{figure}[t!]
\centering
\includegraphics[width=\linewidth]{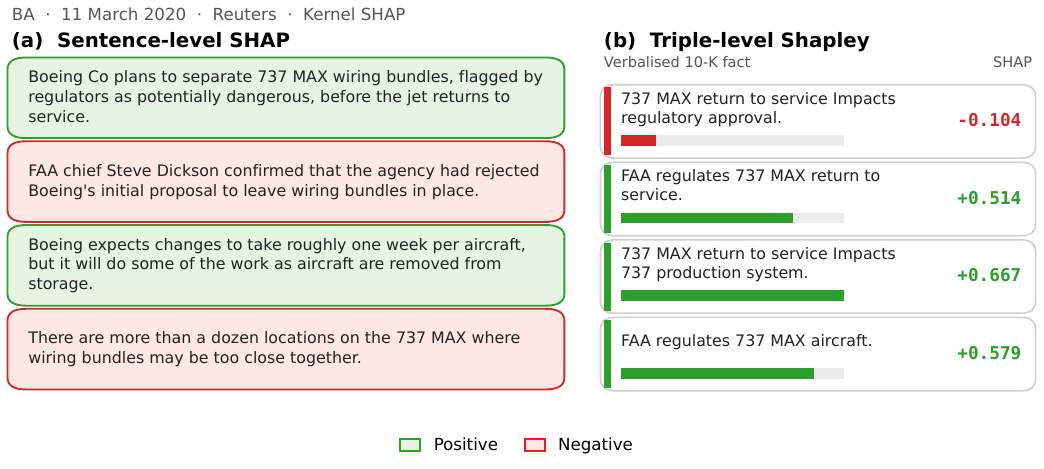}
\caption{Sentence- and triple-level attributions for a top-$k$ triple prediction ($s=+0.790$).
\textbf{(a)}~Selected sentence-level Kernel SHAP attributions.
\textbf{(b)}~Exact Shapley values for the four retrieved 10-K facts relative to the company-anchored prompt.
Green and red indicate positive and negative contributions, respectively.}
\label{fig:explainability}
\end{figure}

We focus on the top-$k$ configuration because it provides explanations at both sentence and knowledge-graph levels. Figure~\ref{fig:explainability} shows sentence-level Kernel SHAP~\cite{kernel_shap} attributions computed while retaining the full retrieved context. Subsets of sentences are removed and rescored, with an empty article assigned a score of zero as the reference. Positive and negative attributions indicate how each sentence within an article shifts the sentiment score. Kernel SHAP is preferred to LIME~\cite{lime} because its weighted regression directly estimates Shapley values and produces additive attributions that satisfy completeness relative to the chosen baseline. In contrast, LIME fits a local surrogate model without a completeness guarantee.

The retrieved triples are attributed separately using exact Shapley values, computed over all $2^k$ subsets while holding the article fixed. The company-anchored prompt without triples provides the reference, so the attributions sum to the difference between the top-$k$ and anchored scores, isolating the effect of the retrieved facts. In this example, the triple attributions sum to $+1.656$, shifting the anchored score from $-0.866$ to the final top-$k$ score of $+0.790$. Together, the sentence- and triple-level explanations show how the model interpreted the article and which retrieved facts changed its prediction, enabling transparent human review of the resulting trading signal.

\section{Conclusion}
We have introduced FinLLM-KG, a novel framework for improving financial sentiment analysis by grounding a frozen causal language model in temporally dated, company-level knowledge. Different from post-training approaches, FinLLM-KG does not update the weights of the underlying LLM, but injects the latest already-public 10-K information at inference time, either as a natural-language company profile or as a small set of relevance-ranked factual relations. This allows any causal LLM to score news against explicit firm-level context rather than in isolation, while the facts underlying each prediction remain inspectable. FinLLM-KG has demonstrated consistent improvements in sentiment-driven portfolio performance over the FinDPO baseline, with the strongest gains obtained from compact, article-specific triples. Further, sentence- and triple-level attributions of the sentiment prediction enhance explainability and support human-in-the-loop review.

\vfill\pagebreak

\section{Compliance with Ethical Standards}
This is a numerical simulation study for which no ethical approval was required.
\bibliographystyle{IEEEbib}
\bibliography{strings,refs}

\end{document}